\documentclass[conference]{IEEEtran}

\IEEEoverridecommandlockouts
\usepackage{cite}
\usepackage{amsmath,amssymb,amsfonts}
\usepackage{algorithmic}
\usepackage{graphicx}
\usepackage{textcomp}
\usepackage{xcolor}
\usepackage{tikz}
\usepackage{adjustbox}
\usepackage{array}
\usepackage{tabularx}
    \newcolumntype{L}{>{\raggedright\arraybackslash}X}
\usetikzlibrary{positioning, backgrounds, calc, arrows}
\def\BibTeX{{\rm B\kern-.05em{\sc i\kern-.025em b}\kern-.08em
    T\kern-.1667em\lower.7ex\hbox{E}\kern-.125emX}}
\graphicspath{ {result_images/} }
\usepackage{hyperref}       
\usepackage{url}            
\usepackage{subcaption}

\usepackage[utf8]{inputenc} 
\usepackage[T1]{fontenc}

\newcommand{\tsn}[1]{{\left\vert\kern-0.25ex\left\vert\kern-0.25ex\left\vert #1 
    \right\vert\kern-0.25ex\right\vert\kern-0.25ex\right\vert}}
\usepackage{xcolor}
\definecolor{darkred}{RGB}{150,0,0}
\definecolor{darkgreen}{RGB}{0,150,0}
\definecolor{darkblue}{RGB}{0,0,200}

\newcommand{\beq}{\begin{equation}}

\newcommand{\eeq}{\end{equation}}

\newcommand{\opnorm}[1]{\left\|#1\right\|}

\newcommand{\abs}[1]{\left|#1\right|}

\def \endprf{\hfill {\vrule height6pt width6pt depth0pt}\medskip}

\begin{document}

\title{De-Biasing Generative Models\\ using Counterfactual Methods\\
{\footnotesize}
\thanks{}
}

\author{\IEEEauthorblockN{Sunay Bhat$^{\dagger}$
\thanks{$^{\dagger}$ \text{Equal Contribution}}
\thanks{\textit{Department of Electrical and Computer Engineering}}
\thanks{\textit{University of California, Los Angeles}}
}
\IEEEauthorblockA{
sunaybhat1@ucla.edu}
\and
\IEEEauthorblockN{Jeffrey Jiang$^{\dagger}$}
\IEEEauthorblockA{
jimmery@ucla.edu}
\and
\IEEEauthorblockN{Omead Pooladzandi$^{\dagger}$}
\IEEEauthorblockA{
opooladz@ucla.edu}
\and
\IEEEauthorblockN{Gregory Pottie$^{\dagger}$}
\IEEEauthorblockA{
pottie@ee.ucla.edu}
}

\newcommand{\algor}{{{Causal Counterfactual Generative Model}}}
\newcommand{\alg}{{\textsc{CCGM}}}

\newcommand{\mean}[2]{\mathbb{E}_{#2}\left[ {#1} \right]}

\maketitle

\begin{abstract}

Variational autoencoders (VAEs) and other generative methods have garnered growing interest not just for their generative properties but also for the ability to dis-entangle a low-dimensional latent variable space. However, few existing generative models take causality into account. We  propose a new decoder based framework named the \algor{} (\alg), which includes a partially trainable causal layer in which a part of a causal model can be learned without significantly impacting reconstruction fidelity. By learning the causal relationships between image semantic labels or tabular variables, we can analyze biases, intervene on the generative model, and simulate new scenarios. Furthermore, by modifying the causal structure, we can generate samples outside the domain of the original training data and use such counterfactual models to de-bias datasets. Thus, datasets with known biases can still be used to train the causal generative model and learn the causal relationships, but we can produce de-biased datasets on the generative side. 
Our proposed method combines a causal latent space VAE model with specific modification to emphasize causal fidelity, enabling finer control over the causal layer and the ability to learn a robust intervention framework. We explore how better disentanglement of causal learning and encoding/decoding generates higher  causal intervention quality. We also compare our model against similar research to demonstrate the need for explicit generative de-biasing beyond interventions. Our initial experiments show that our model can generate images and tabular data with high fidelity to the causal framework and accommodate explicit de-biasing to ignore undesired relationships in the causal data compared to the baseline.

\end{abstract}

\section{Introduction}

In many fields such as medicine and economics, an explainable model, in particular a causal model, is needed to elicit the effectiveness of interventions. This process makes diligent use of prior knowledge, usually in a structural causal model (SCM) that instantiates unidirectional relationships between the variables using a Directed Acyclic Graph (DAG) \cite{pearl-paper}. The confidence needed in a causal model needs to be much higher than in a statistical model as one needs to instantiate beliefs that are invariant and exist outside the domain of the data. Traditionally, this knowledge comes from experimentally derived results, or domain experts with experimental level knowledge. As such, there is a strong interest in the deep learning community to integrate causal methods and information more directly with traditional deep learning architectures. Although recent results show progress in causal deep learning, most methods  focus on either causal discovery or the use of prior causal information alone \cite{causal-rl,causalgan,dag-gnn}.

Generative models have been crucial to solving many problems in modern machine learning \cite{vanilla-vae}. Since the VAE's inception, many have found that the disentanglement of latent spaces can lead to better performance in generalizability and fine-tuned control over disentangled features. In addition, many techniques have been proposed in recent years as to how to improve disentanglement, largely based on factorization and independence techniques \cite{betavae, beta-tcvae}.

Recently, an effective approach that blends the space of causal models with generative neural networks was displayed with the CausalVAE, which allowed the decoder to learn a \textit{causally} disentangled representation of latent space variables. One of the key contributions in that paper was the inclusion of a \textit{Causal Layer}. Most impressively, the CausalVAE enforced a causal structure on generating images to noticeably disentangle intentionally dependent latent variables via the use of a causal layer. This causal layer's disentanglement allows the CausalVAE to generate causal interventions. Specifically, when intervening on endogenous variables, the CausalVAE is able to generate images that are outside the normal bounds of the training dataset, as the intervention does not affect the exogenous variables \cite{causalvae}. 

Here, we combine the ideas of counterfactual causal reasoning and generative modeling by focusing on the causal layer of the CausalVAE. We modify the objective into learning a more refined, isolated causal structure that the latent space must go through, which we call the \algor. This allows us to expand the use of the causal layer to more than just single interventions, to also to hypothesize and synthesize datasets of counterfactual causal models in interesting and useful ways. 

\section{Related Work}
Causal discovery has increasingly been the focus of deep learning methods which seek to reduce the combinatorial complexity of brute force searches for causal models from observational data. Progress in DAG search using continuously differentiable loss functions and reinforcement learning for score functions has started to integrate deep learning methods with causal discovery and identification \cite{dag-no-tears}\cite{causal-rl}. 

Building on initial deep causal discovery, causal generative models learn or use causal information for generating data and interventions. CausalGAN is a generative model that learns a prior Structural Causal Model (SCM) for images and label spaces and demonstrates how interventions in the latent space can generate causally intervened images \cite{causalgan}. DAG-GNN uses graph neural networks with a VAE architecture to extend causal discovery methods to more use-cases \cite{dag-gnn}. CausalVAE uses a causal layer in the middle of a VAE architecture to learn an implicit causal model that can also generate unseen images with latent space interventions \cite{causalvae}. Causal discovery with generative models capitalize on recent work in disentanglement to ensure the latent space has the necessary variable structure for causal identification \cite{betavae}. Finally, causal generative models have been used to address the issue of fair or ``de-biased" data sets such as DECAF, a causally aware GAN architecture applied explicitly to tabular data \cite{decaf}. 

When causal models are known or hypothesized to contain measured confounders, statistical adjustment techniques have long been used to estimate causal effects when the structure is known or identifiable. Inverse Propensity Score Weighting (IPW), or advanced methods like Augmented IPW provide robust or doubly-robust ways to adjust for confounding bias \cite{glynnquinnAIPW}.

\section{Background}

\subsection{Counterfactuals and Interventions}

The SCM literature has long explored the benefits of interventions and counterfactual modeling once a causal model is known. Pearl introduces interventions using `do-calculus' or the explicit setting of a variable to a specific value and calculating the resulting outputs \cite{pearl-paper}. In Figure \ref{fig:interventions} below, we introduce a 4-variable DAG with two exogenous and two endogenous variables. An intervention on the right shows how this is effectively breaking the parent nodes into the variable being intervened on, and explicitly setting it to a desired value ($x$), written using do-calculus notation $do(x)$. This operation allows us to directly fix the value of a latent variable and asymmetrically propagate its value to other variables. Intervened parents should have their adjusted values  impact child nodes, but intervened children should not adjust parent values. 


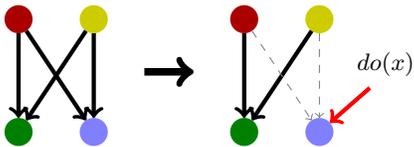
\begin{figure}[ht]
    \centering
    \begin{tikzpicture}

    \node[draw, shape=circle, color={rgb:black,1;green,1}, fill={rgb:black,1;green,1}, very thick, minimum size = 0.02cm] at (0, 0) (x3) {};
    \node[draw, shape=circle, color={rgb:black,1;red,2}, fill={rgb:black,1;red,2}, very thick, minimum size = 0.02cm] at (0, 1.5) (x1) {};
    \node[draw, shape=circle, color={rgb:black,1;yellow,4}, fill={rgb:black,1;yellow,4}, very thick, minimum size = 0.02cm] at (1, 1.5) (x2) {};
    \node[draw, shape=circle, color=blue!50, fill=blue!50, very thick, minimum size = 0.02cm] at (1, 0) (x4) {};
    

    \draw[ultra thick, ->] (x1) -- (x3);
    \draw[ultra thick, ->] (x1) -- (x4);
    \draw[ultra thick, ->] (x2) -- (x3);
    \draw[ultra thick, ->] (x2) -- (x4);
    
    \node[draw, shape=circle, color={rgb:black,1;green,1}, fill={rgb:black,1;green,1}, very thick, minimum size = 0.02cm] at (3, 0) (x3_2) {};
    \node[draw, shape=circle, color={rgb:black,1;red,2}, fill={rgb:black,1;red,2}, very thick, minimum size = 0.02cm] at (3, 1.5) (x1_2) {};
    \node[draw, shape=circle, color={rgb:black,1;yellow,4}, fill={rgb:black,1;yellow,4}, very thick, minimum size = 0.02cm] at (4, 1.5) (x2_2) {};
    \node[draw, shape=circle, color=blue!50, fill=blue!50, very thick, minimum size = 0.02cm] at (4, 0) (x4_2) {};
    \node[shape=circle] at (4.8, 0.7) (inter) {};
    \node [text width=1cm] (0) at (5,0.9) {\small $do(x)$};
        \node [text width=1cm] (0) at (-0.6,0.9) {};
    
    
    \node[shape=circle] at (1.5, 0.8) (arrow_1) {};
    \node[shape=circle] at (2.5, 0.8) (arrow_2) {};
    
    
    \draw[ultra thick, ->] (x1_2) -- (x3_2);
    \draw[ultra thick, ->] (x2_2) -- (x3_2);
    \draw[dashed, ->,gray] (x1_2) -- (x4_2);
    \draw[dashed, ->,gray] (x2_2) -- (x4_2);
    \draw[ultra thick, ->, line width=1mm] (arrow_1) -- (arrow_2);
    \draw[ultra thick, ->,red] (inter) -- (x4_2);
    
\end{tikzpicture}
    \caption{Example of a DAG on the left and a mutated counterfactual model on the right with an intervention setting the target variable to an explicit value $x$.}
    \label{fig:interventions}
\end{figure}

\subsection{Counterfactual Models}

\begin{figure}[ht]
    \centering
    \begin{tikzpicture}

    \node[draw, shape=circle, color={rgb:black,1;green,1}, fill={rgb:black,1;green,1}, very thick, minimum size = 0.02cm] at (0, 0) (x3) {};
    \node[draw, shape=circle, color={rgb:black,1;red,2}, fill={rgb:black,1;red,2}, very thick, minimum size = 0.02cm] at (0, 1.5) (x1) {};
    \node[draw, shape=circle, color={rgb:black,1;yellow,4}, fill={rgb:black,1;yellow,4}, very thick, minimum size = 0.02cm] at (1, 1.5) (x2) {};
    \node[draw, shape=circle, color={rgb:black,1;blue,2}, fill={rgb:black,1;blue,2}, very thick, minimum size = 0.02cm] at (1, 0) (x4) {};
    
    \node [text width=1cm] (0) at (0.4, 1.9) {\small $\theta$};
    \node [text width=1cm] (0) at (1.2,1.9) {\small $x_{sun}$};
    \node [text width=1cm] (0) at (-0.4,-0.4) {\small $w_{shadow}$};
    \node [text width=1cm] (0) at (1.2,-0.4) {\small $x_{shadow}$};

    \draw[ultra thick, ->] (x1) -- (x3);
    \draw[ultra thick, ->] (x1) -- (x4);
    \draw[ultra thick, ->] (x2) -- (x3);
    \draw[ultra thick, ->] (x2) -- (x4);
    
    \node[draw, shape=circle, color={rgb:black,1;green,1}, fill={rgb:black,1;green,1}, very thick, minimum size = 0.02cm] at (3, 0) (x3_2) {};
    \node[draw, shape=circle, color={rgb:black,1;red,2}, fill={rgb:black,1;red,2}, very thick, minimum size = 0.02cm] at (3, 1.5) (x1_2) {};
    \node[draw, shape=circle, color={rgb:black,1;yellow,4}, fill={rgb:black,1;yellow,4}, very thick, minimum size = 0.02cm] at (4, 1.5) (x2_2) {};
    \node[draw, shape=circle, color={rgb:black,1;blue,2}, fill={rgb:black,1;blue,2}, very thick, minimum size = 0.02cm] at (4, 0) (x4_2) {};
    
    \node [text width=1cm] (0) at (3.4, 1.9) {\small $\theta$};
    \node [text width=1cm] (0) at (4.2,1.9) {\small $x_{sun}$};
    \node [text width=1cm] (0) at (2.6,-0.4) {\small $w_{shadow}$};
    \node [text width=1cm] (0) at (4.2,-0.4) {\small $x_{shadow}$};
        
    \node[shape=circle] at (4.8, 0.7) (inter) {};
    \node [text width=1cm] (0) at (5,0.1) {\small $x=f(\theta)$};
    
    \node[shape=circle] at (1.5, 0.8) (arrow_1) {};
    \node[shape=circle] at (2.5, 0.8) (arrow_2) {};
    
    
    \draw[ultra thick, ->] (x1_2) -- (x3_2);
    \draw[ultra thick, ->] (x2_2) -- (x3_2);
    \draw[ultra thick, ->,color={rgb:black,1;green,1}] (x1_2) -- (x4_2);
    \draw[dashed, ->,gray] (x2_2) -- (x4_2);
    \draw[ultra thick, ->, line width=1mm] (arrow_1) -- (arrow_2);

\end{tikzpicture}
    \caption{Example of a counterfactual model in which a single path is removed to simulate a new distribution of generated data. }
    \label{fig:counter_model}
\end{figure}
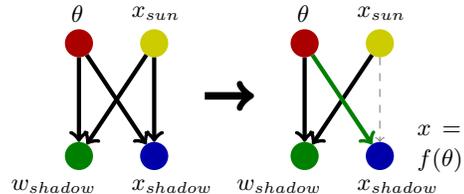

Extending from the idea of interventions on instances of data, we define counterfactual models as a new model formed by removing a path deemed undesirable or a source of bias as seen in Figure \ref{fig:counter_model}. This could be a known bias present in the data generating process, or a desire to envision a new data distribution outside the training dataset with a specific graphical modification. Notice, unlike an intervention as in Figure \ref{fig:interventions}, the target variable need not be set explicitly but still is a function of the other parent variables. This allows a data distribution to be generated in which the target is still a function of the remaining parent nodes, possibly simulating a ``de-biased" or counterfactually constructed dataset, as opposed to explicit instantiations of the intervened variable. 

\subsection{Constructing a Causal Generative Model}

Following the classic VAE model, given inputs $\mathbf{x}$, we encode into a latent space $\mathbf{z}$ with distribution $q_\phi$ where we have priors given by $p(\cdot)$ \cite{vanilla-vae}. 
\begin{equation} \label{eq:elbo}
    \text{ELBO} = \mean{\mean{\log p_\theta(\mathbf{x} \vert \mathbf{z})}{\mathbf{z} \sim q_\phi}- \mathcal{D}( q_\phi(\mathbf{z} \vert \mathbf{x}) \| p_\theta(\mathbf{z}))}{q_{\mathcal{X}}}
\end{equation} 

In \cite{causalvae}, the causal layer is described as a noisy linear SCM:  
\begin{equation} \label{eq:cvae-causal}
    \mathbf{z} = \mathbf{A}^T\mathbf{z} + \pmb{\epsilon}
\end{equation}
which finds some causal structure of the latent space variables $\mathbf{z}$ with respect to a matrix $\mathbf{A}$. By itself, $\mathbf{A}$ functions as the closest linear approximator for the causal relationships in the latent space of $\mathbf{z}$. 

A non-linear mask can be applied to the causal layer so that it can more accurately estimate non-linear situations as well. Suppose $\mathbf{A}$ is composed of column vectors $\mathbf{A}_i$. For each latent space concept $i$, define a non-linear function $g_i : \mathbb{R}^n \rightarrow \mathbb{R}$ and modify equation \eqref{eq:cvae-causal} such that 
\begin{equation} \label{eq:cvae-nonlin-causal}
    \mathbf{z}_i= g_i(\mathbf{A}_i \circ \mathbf{z}) + \pmb \epsilon
\end{equation}
where $\circ$ is the Hadamard product. In this formulation, the view of $\mathbf{A}$ changes from one of function estimation to one of adjacency. That is, if $\mathbf{A}$ is viewed as a binary adjacency matrix, the $g_i$ functions take the responsibility of reconstructing $\mathbf{z}$ given only the the parents, dictated by $\mathbf{A}_i \circ \mathbf{z}$. In the simplest case, if $g_i(\mathbf{v}) = \sum_j v_j$, the summation of all the values of $\mathbf{v}$, then Equation \eqref{eq:cvae-nonlin-causal} degenerates back to Equation \eqref{eq:cvae-causal} \cite{gnet}. 

Including the causal layer introduces many auxiliary loss functions that we mostly adopt \cite{causalvae}. First is a label loss \eqref{eq:u-loss}, where the adjacency matrix $\mathbf{A}$ should also apply to the labels $\mathbf{u}$. This loss is used in pre-training in its linear form to learn a form of $\mathbf{A}$ prior to learning the encoder and decoders. After pre-training, we apply a nonlinear mask $f_i$ that functions similarly to $g_i$, but operates on the label space directly, but with the same $\mathbf{A}$. 
\begin{equation} \label{eq:u-loss}
    \ell_u = \mean{ \sum_{i=1}^n \opnorm{u_i - f_i(\mathbf{A}_i \circ \mathbf{u})}^2}{q_\mathcal{X}}
\end{equation}

The latent loss tries to enforce the SCM, described by Equation \eqref{eq:cvae-nonlin-causal}. 
\begin{equation} \label{eq:latent-loss}
    \ell_z = \mean{\sum_{i=1}^n \opnorm{z_i - g_i(\mathbf{A}_i \circ \mathbf{z})}^2}{\mathbf{z} \sim q_\phi}
\end{equation}

Further enforcing the label spaces, we can define a prior $p(\mathbf{z} \vert \mathbf{u})$. We use the same conventions as in \cite{causalvae} and say that 
$$p(\mathbf{z} \vert \mathbf{u}) \sim \mathcal{N}(\mathbf{u}_n, \mathbf{I}) $$
where $\mathbf{u}_n \in [-1,1]$ are normalized label values. This translates to an additional KL-loss. 

Finally, we apply the continuous differentiable loss function \eqref{eq:dag-loss} and apply a scheduling technique to enforce the DAG \cite{dag-no-tears,dag-gnn}. The main use is that $\mathbf{G}$ is a DAG if and only if 
\begin{equation} \label{eq:dag-loss}
    H(\mathbf{G}) := tr\left[ \left( \mathbf{I} + \mathbf{G} \circ \mathbf{G} \right)^n \right] - n = 0
\end{equation}

The scheduling is done via the augmented Lagrangian
\begin{equation}
    \ell_h = \lambda H(\mathbf{G}) + \frac{c}{2} \abs{h(\mathbf{G})}^2
\end{equation}
where at the end of every epoch, the scheduling update is 
\begin{align} \label{eq:dag-sched}
    \lambda_{t+1} &= \lambda_t + c_t H(\mathbf{G}_t) \\
    c_{t+1} &= \begin{cases}
    \eta c_t & \abs{H(\mathbf{G}_t)} > \gamma \abs{H(\mathbf{G}_{t-1})} \\
    c_t & \text{else}
    \end{cases} \nonumber
\end{align}
where we set $\eta = 2$ and $\gamma = 0.9$. 

\subsection{Causal Estimation}

There are numerous ways to estimate a causal effect once the model has been identified. Perhaps the most common and simplest is the Average Treatment Effect ($\widehat{ATE}$), which is simply the difference of means between a population ($index = i$) treated and untreated group, assuming a binary intervention variable ($D$), and an outcome variable ($Y$) as in equation \eqref{eq:naive-ate}. 
\begin{align}
    \widehat{ATE} = \mathbb{E}[Y_{i}|D_i = 1]-\mathbb{E}[Y_{i}|D_i = 0]
    \label{eq:naive-ate}
\end{align}

This naïve method does not consider any confounding variables. One common way to adjust for such confounding bias is to use propensity scores ($\hat \pi (X_i)$), which is a model for how likely a sample is to receive the treatment based on the measured covariate factors. The inverse of the propensity score can then be used to weight each sample as in equation \eqref{eq:ipw-ate} and thus adjust for the bias of any measured confounders. 
\begin{align}
    \widehat{ATE}_{IPW} = \frac{1}{N} \sum_{i=1}^N \left[\frac{D_iY_i}{\hat \pi (X_i)} - \frac{(1-D_i)Y_i}{1-\hat \pi (X_i)}\right]
    \label{eq:ipw-ate}
\end{align}

Finally, more recent developments in double-robust methods specify both an outcome model and an exposure/propensity score model which can provide accurate estimation if either one of the models is misspecified. Augmented IPW (AIPW) is a specific method that extends IPW below with a set of outcome models estimating the outcome variable as a function of the intervention and all covariates as introduced in \cite{glynnquinnAIPW}.


\section{Problem Setting}

\subsection{Sun Pendulum Image Dataset}

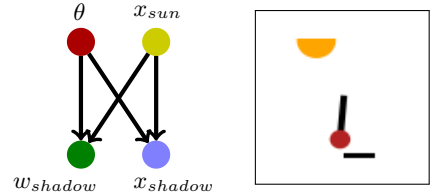
\begin{figure}[ht]
    \centering
    \hspace{0.06\textwidth}
    \begin{subfigure}{0.15\textwidth}
        \begin{tikzpicture}

    \node[draw, shape=circle, color={rgb:black,1;green,1}, fill={rgb:black,1;green,1}, very thick, minimum size = 0.02cm] at (0, 0) (x3) {};
    \node[draw, shape=circle, color={rgb:black,1;red,2}, fill={rgb:black,1;red,2}, very thick, minimum size = 0.02cm] at (0, 1.5) (x1) {};
    \node[draw, shape=circle, color={rgb:black,1;yellow,4}, fill={rgb:black,1;yellow,4}, very thick, minimum size = 0.02cm] at (1, 1.5) (x2) {};
    \node[draw, shape=circle, color=blue!50, fill=blue!50, very thick, minimum size = 0.02cm] at (1, 0) (x4) {};
    
    \node [text width=1cm] (0) at (0.4, 1.9) {\small $\theta$};
    \node [text width=1cm] (0) at (1.2,1.9) {\small $x_{sun}$};
    \node [text width=1cm] (0) at (-0.4,-0.4) {\small $w_{shadow}$};
    \node [text width=1cm] (0) at (1.2,-0.4) {\small $x_{shadow}$};

    \draw[ultra thick, ->] (x1) -- (x3);
    \draw[ultra thick, ->] (x1) -- (x4);
    \draw[ultra thick, ->] (x2) -- (x3);
    \draw[ultra thick, ->] (x2) -- (x4);
    
\end{tikzpicture}
    \end{subfigure}
    \hspace{0.02\textwidth}
    \begin{subfigure}{0.23\textwidth}
        \fbox{\includegraphics[width=.5\linewidth]{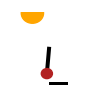}}
    \end{subfigure}
    \caption{Pendulum toy image dataset DAG and example image}
    \label{fig:pendulum}
\end{figure}

A toy pendulum image dataset is introduced in \cite{causalvae}. This dataset is generated by sweeping sun positions ($x_{sun}$) and pendulum angles ($\theta$) to produce realistic shadow width ($w_{shadow}$) and shadow locations ($x_{shadow}$) from deterministic non-linear functions. Figure \ref{fig:pendulum} shows the DAG for this model and an example generated image, in which the sun and pendulum variables are exogenous, and the shadow variables are endogenous. Thus any causal model will learn to reconstruct the shadow variables from the sun and pendulum variables. Such relationships in observational studies are often invertible as correlation has no directionality.  Thus, without causal disentangling, an intervention on shadow position would likely adjust the sun position to match. 

This dataset is used to demonstrate causal generative model quality through reconstruction fidelity as well as causal learning by intervening on parent and child nodes, showing interventions only propagate forward from parents to children and not vice-versa \cite{causalvae}. 

\subsection{Tabular National Study of Learning Mindsets Data}

\begin{figure}[ht]
    \centering
    \begin{tikzpicture}

    \node[draw, shape=circle, color={rgb:black,1;green,1}, fill={rgb:black,1;green,1}, very thick, minimum size = 0.02cm] at (0, 0) (x3) {};
    \node[draw, shape=circle, color={rgb:black,1;red,2}, fill={rgb:black,1;red,2}, very thick, minimum size = 0.02cm] at (0, 1.5) (x1) {};
    \node[draw, shape=circle, color={rgb:black,1;yellow,4}, fill={rgb:black,1;yellow,4}, very thick, minimum size = 0.02cm] at (1, 1.5) (x2) {};
    \node[draw, shape=circle, color=blue!50, fill=blue!50, very thick, minimum size = 0.02cm] at (1, 0) (x4) {};
    
    \node [] (0) at (-0.7, 1.9) {\scriptsize School Mindset};
    \node [] (0) at (1.9,1.9) {\scriptsize Success Expectation};
    \node [] (0) at (-0.6,-0.4) {\scriptsize Intervention};
    \node [] (0) at (1.9,-0.4) {\scriptsize Achievement Score};

    \draw[ultra thick, ->] (x1) -- (x3);
    \draw[ultra thick, ->] (x1) -- (x4);
    \draw[ultra thick, ->] (x2) -- (x3);
    \draw[ultra thick, ->] (x2) -- (x4);
    
\end{tikzpicture}
    \caption{School Mindset DAG}
    \label{fig: mindset_DAG}
\end{figure}
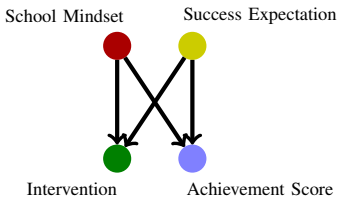

To analyze our methods in a tabular setting, we use a simulated dataset based on The National Study of Learning Mindsets \cite{mindsets}. This was a randomized study conducted in U.S. public high schools, the purpose of which was to evaluate the impact of a nudge-like intervention designed to instill students with a growth mindset on student achievement. We use a simulated subset of the data based on a model fit to the statistics of the original dataset (the actual dataset was not publicly released). The study includes measured outcomes via an achievement score, a binary treatment of a growth mindset educational intervention (not to be confused with a causal intervention), and 11 other potential confounding factors that could be parents of both the treatment and outcome. We select two of these confounding variables: an average measure of the fixed mindset at each student's school (inversely correlated with achievement and educational intervention) and the students' self-reported expectations of their own success (positively correlated with achievement and educational intervention). The full correlations between all four variables can be seen in Table \ref{table:corr_data}. Thus we maintain a hypothesized DAG structure as in Figure \ref{fig: mindset_DAG} identical to the pendulum model. Note that our interest is in regenerating the dataset with the treatment and targets as functions of the confounders, so we do not learn the effect of the intervention on the outcome. We will use our methods to generate datasets in which we can estimate the ATE to estimate our causal effect using a simple difference of means.

The intuitive belief is that a naïve estimate of the ATE, calculated as the difference of means as in equation \eqref{eq:naive-ate}, would contain a positive bias due to the confounding variable of a student's own expectation. Students with higher expectations are more likely to participate in the growth mindset course (self-selection bias) but are also likely to have higher achievement anyway. Statistical adjustment techniques, such as Inverse Propensity Weighting (IPW), attempt to control for such confounders by measuring and weighting the effect based on the propensity to be treated. We will use such methods as a baseline for comparison, as we will first generate a dataset approximating the existing data distribution while learning some causal features. We will then employ a counterfactual model removing a confounding link and demonstrate a simulated dataset in which the naïve ATE aligns with the ATE measure using the statistical adjustment methods. 

\begin{table}[!t]
\renewcommand{\arraystretch}{1.3}
\newcommand*\rot{\rotatebox{0}}
\caption{Correlation of Mindset Variables}
\label{table_example}
\centering
\begin{tabular}{c|cccc}
\hline {} & \rot{SM} & \rot{SE} & \rot{D} & \rot{Y} \\
\hline
    \scriptsize School Mindset (SM)    &        1&       -0.054 &     -0.046 &          -0.111 \\
    \scriptsize Success Expectation (SE)    &       -0.054 &        1 &      0.059 &           0.439 \\
    \scriptsize Intervention (D)      &       -0.046 &        0.059 &      1 &           0.221 \\
    \scriptsize Achievement Score (Y) &       -0.111 &        0.439 &      0.221 &           1 \\
\hline
\end{tabular}
\label{table:corr_data}\vspace{-3mm}
\end{table}
\section{Causal Counterfactual Generative Model}
We start from many of the same concepts as the original CausalVAE but begin by changing the enforced structure of the causal layer, allowing us to make direct modifications to the layer after training.

\subsection{Limits of the CausalVAE for Counterfactuals}

The causal layer in \cite{causalvae} has a purpose of passing some causal information about the latent space through from parents to children. However, there is a fundamental difference in how we would like to interpret our problem. Reiterating \eqref{eq:cvae-causal}, 
$$\mathbf{z} = \mathbf{A}^T\mathbf{z} + \pmb{\epsilon}$$
Whatever causal structure is learned by $\mathbf{A}$, there will always be a ``leakage" of information via $\pmb{\epsilon}$. This $\pmb \epsilon$ can be viewed as the output of a vanilla VAE, meaning that theoretically it can contain contribute everything for image generation. This leakage informs $\mathbf{z}$ without passing through the causal layer, so it weakens the need for $\mathbf{A}$ to learn all the causal structure of the problem. In the image space, this leakage of information improves generation and reconstruction and hence is desirable. However, it does not align with our objective of finding a good underlying causal structure. In the most extreme case, we could, in theory, find $\mathbf{A} = \mathbf{0}$, which is still a valid DAG. In this case, no remaining causal information remains in the layer and the entire CausalVAE reverts back to a normal $\beta$-VAE.

\subsection{Envisioning Bias-Free Models with CausalVAE}

Here, we introduce \alg{} as a modified and extended version of the CausalVAE, allowing for counterfactual models. In particular, we can directly manipulate the causal layer so that undesirable causal links learned from the data can be broken.  

In \alg{}, the encoder directly generates the output $\mathbf{z}$, which is enforced to be standard-normally distributed. We pass this through our causal layer as one final \textit{mutatable} bottleneck
\begin{equation} \label{eq:lin-causal}
    \mathbf{z} = \mathbf{A}^T \mathbf{z}
\end{equation}
That is, it instantiates a linear SCM. One main distinction is that we solidify the structure of $\mathbf{A}$ by having exogenous and endogenous priors. This way, $\mathbf{A}$ can be split into a DAG term and a diagonal term: 
\begin{equation}
    \mathbf{A} = \underbrace{\mathbf{G}}_{\text{DAG}} + \underbrace{\mathbf{B}}_{\text{diag.}}
\end{equation}
where $\mathbf{B}$ has 1 on the diagonal for exogenous variables and 0 if endogenous. This ensures that the trivial solution where $\mathbf{A} = \mathbf{I}$ is never learned and enforces a causal relationship from the exogenous variables to the endogenous variables.

Similarly, we add the non-linear mask to the causal layer just as in equation \eqref{eq:cvae-nonlin-causal}, but dropping the leakage. 
\begin{equation} \label{eq:nonlin-causal}
    \mathbf{z}_i= g_i(\mathbf{A}_i \circ \mathbf{z})
\end{equation}
When separating adjacency and estimation, we necessarily want to have a pre-training step for 5 epochs, where we train $\mathbf{A}$ to recognize the adjacency of the labels before applying the non-linear mask. After the pre-training, we apply training on both the $\mathbf{A}$ matrix and the non-linear mask, but there should be fewer changes as the mask should take care of the function approximations.

The ultimate goal of our work is to propose counterfactual causal models by directly manipulating this $\mathbf{A}$ matrix. The framework proposed by \cite{causalvae} requires one to retrain the entire model to generate a counterfactual $\mathbf{A}$ while fixing a path in the graph to zero, as their intervention method does not deal with the leakage. This is an expensive task in both time and computing power, making it unscalable for larger $\mathbf{A}$'s. Our method allows us to generate data about a hypothesized counterfactual space directly by breaking links in the causal graph, without the need to retrain the neural network.

\subsection{General Structure of \alg{}}

While one could work with image-to-image VAEs, in our examples, we leverage as much tabular data as we can to reduce computational needs. In the pendulum example, we know the labels can be used as a perfect reconstruction of the data and so the labels that are provided act as at least a perfect bottleneck, containing more information than needed, in the reconstruction of images. 

Furthermore, the label-to-label structure can be used as a pre-training step in determining a causal matrix. It then becomes a natural extension to apply the \alg{} to tabular data. We no longer require a VAE setup, although we preserve the mild non-linear networks which allow for more complex causal functionality. Our experiments section will show a \alg{} capable of generating tabular data with a reasonable representative distribution and a bias removed distribution. Note that noisy tabular data with hypothesized causal models (no known ground truth model or guarantee of endogenous/exogenous priors) present a new set of identification and estimation challenges.

\section{Experiments}

In this section, we evaluate the effectiveness of causal generative models on tabular and image datasets, by answering the following questions: (1) how does the performance of \alg{} compare to the state of the art methods in reconstruction and causal logic; (2) how effective is \alg{} for eliminating biases in image and and tabular datasets; and (3) how \alg{} generates counterfactual models without extra training allowing for diverse and flexible data-generation. We compare the performance of \alg{} and CausalVAE to generate counterfactual samples from a fixed causal model \cite{causalvae}. We further compare \alg{} to advanced statistical adjustment methods for generating ``de-biased" datasets vs. controlling biases statistically. 

\subsection{CausalVAE}
\begin{figure}
    \centering
    \includegraphics[width=0.4\textwidth,trim={1mm 6mm 1mm 1mm},clip]{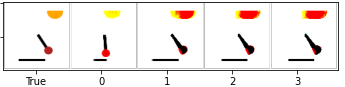} \vspace{1mm}
    \includegraphics[width=0.4\textwidth,trim={1mm 1mm 1mm 1mm},clip]{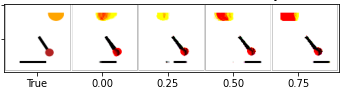}
    \caption{A sweep of the pendulum angle (top) and sun position (bottom) in the latent space for CausalVAE. Values are chosen to attempt to see changes. Outside of the 0 intervention, other interventions do not seem to make sense, especially with the shadows. }
    \label{fig:orig_sweeps}\vspace{0mm}
\end{figure}

Our experiments with the standard CausalVAE found that the model could handle interventions on specific latent space data, meaning that its decoder could causally disentangle some of the concepts. However, non-zero interventions did not appear to be working as intended. Figure \ref{fig:orig_sweeps} shows a sweep of interventions on the pendulum and sun position data, respectively, on the same image. Notice that the first intervention, corresponding to 0, works as intended. However, the pendulum does not change outside the 0 value, while the sun changes somewhat in an expected fashion, but the shadow does not respond. 

\begin{figure}
    \centering
    \includegraphics[width=0.2\textwidth]{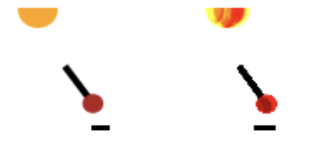}
    \caption{Result which still shows effect propagation (shadow moves) after removing the path from sun location to shadow location in CausalVAE method}
    \label{fig:Baseline_no_change}
    \vspace{-3mm}
\end{figure}

Furthermore, we noticed little to no change in the results of certain interventions when generating counterfactual models, such as in Figure \ref{fig:Baseline_no_change} below where a post training removal on the path from sun location to shadow position did not remove the effect propagation. These findings reflect that the CausalVAE was not designed to learn the full causal structure due to the leakage in $\pmb{\epsilon}$. 

\subsection{Label-to-Label}
Our initial experiments pertain to the label-to-label space, where we have four parameters (labels) that provide perfect information for image reconstruction.

We start with a set of labels $\mathbf{u} \in \mathbb{R}^n$, where $n=4$. These four labels correspond to the pendulum angle $\theta$, the sun position $x_{sun}$, the shadow width $w_{shadow}$, and the shadow position $x_{shadow}$, respectively such that  $\mathbf{u} = [\theta,x_{sun},w_{shadow},x_{shadow}]^T $. Then, $\mathbf{u}$ passes through an encoder to generate $\mathbf{z} \in \mathbb{R}^n$, where we enforce the prior of $p(z) \sim \mathcal{N}(0, \mathbf{I})$. This step allows us to sample new labels from the latent space drawn from a Gaussian distribution, as we see in the classic VAE \cite{vanilla-vae}. 
Now, we subject $\mathbf{z}$ to the learned causal layer. Based on our designation of $\mathbf{u}$, we set $diag(\mathbf{A}) = [1, 1, 0, 0]$, representing the exogenous and endogenous variables of $\mathbf{u}$. Equation \ref{eq:nonlin-causal} is applied to to $\mathbf{z}$, and the information of $\mathbf{z}$ should be preserved through the causal layer, even though the exact information of the endogenous variables is intentionally dropped. 

Finally, this reconstructed latent space vector $\hat{\mathbf{z}}$ is passed into a decoder to reconstruct the original labels $\hat{\mathbf{u}}$. For consistency of visualization and easy of human understanding, we pass these labels into a separate image generator to create all visualized images.

\textit{\alg{} Generates Clean Label-to-Label Interventions.}
Since there are few parameters in the label-to-label space we are able to generate clean counterfactual models as well as interventions.



\begin{figure}
    \centering
    \includegraphics[width=0.4\textwidth,trim={1mm 8.7mm 1mm 3mm},clip]{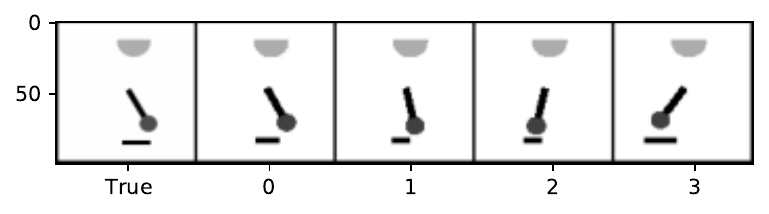}
    \includegraphics[width=0.4\textwidth,trim={1mm 3mm 1mm 3mm},clip]{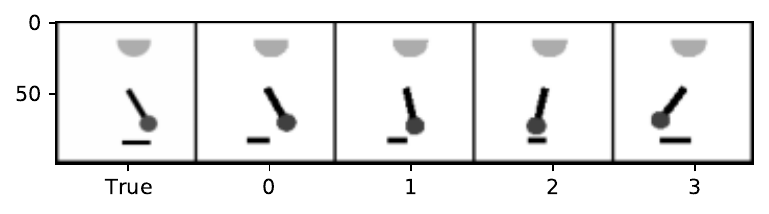}
    \caption{Label-to-label (Top) Image response to a sweep of the pendulum angle. Notice that for all interventions, the shadow responds to the pendulum. (Bottom) Image response after shadow position is debiased from pendulum angle. In particular pay attention to the right-most image. While subtle, the shadow positions between the top and bottom image are very noticeable. A quick scan from left to right on all of the intervened images suggests that the midpoint of the shadow remains constant throughout all of the swept images. However, it is worth noting that the shadow width still responds as if the shadow had moved to its location. }
    \label{fig:l2l_sweep_pendulum}
    \vspace{-3mm}
\end{figure}

\begin{figure}
    \centering
    \includegraphics[width=0.4\textwidth,trim={1mm 8.5mm 1mm 3mm},clip]{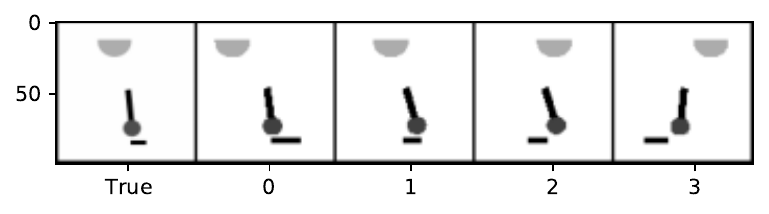}
    \includegraphics[width=0.4\textwidth,trim={1mm 3mm 1mm 3.5mm},clip]{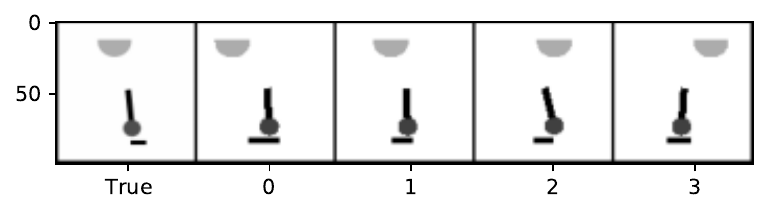}
    \caption{Label-to-label (Top) Image response to a sweep of sun positions. Again, notice that the shadow responds to the sun's position. (Bottom) In this case, the changes are more noticeable in that the shadow position remains constant throughout the row and is very different than the expected locations given the sun.}
    \label{fig:l2l_sweep_sun}\vspace{-4mm}
\end{figure}

Our primary results for label-to-label is shown in Figures \ref{fig:l2l_sweep_pendulum} and \ref{fig:l2l_sweep_sun}. The top row of both figures show interventional sweeps on both $\theta$ and $x_{sun}$, respectively. We take a true image and apply a range of interventions sampled from the range of the resultant sampling distribution ($\mathcal{N}(0,1)$) to generate counterfactual samples. Interventions on exogenous variables shows a response in the shadow variables, but the other exogenous variable should stay constant. 

Then, we apply the ideas of a counterfactual model. Instead of doing interventions on specific values, we break the link of $x_{shadow}$ with $\theta$ and $x_{sun}$ in $\mathbf{A}$, respectively. Afterward, if we do the same interventions, the shadow position no longer responds to that intervention. While some of the results can be subtle, in Figure \ref{fig:l2l_sweep_pendulum}, the final image shows a noticeable difference in position before and after the counterfactual model and in Figure \ref{fig:l2l_sweep_sun}, the first and last interventions both show differences. The connection to shadow width remains, and so the shadow width still responds to the swept variable.

\subsection{Label-to-Image}
We then consider the more challenging problem of generating an entire image from the label information. Thus, we propose a label-to-image generative model based on the decoder of a VAE to use the disentanglement granted by the Causal Layer. We can use a pre-trained version of the $\mathbf{A}$ matrix coming from the label-to-label VAE to start our training of the causal generative model. For the sake of computational power, we keep the dataset in grayscale to reduce the image size by at least a factor of 3, but the physics aspects are still present. 

\begin{figure}
    \centering
    \includegraphics[width=0.4\textwidth,trim={1mm 3mm 1mm 3mm},clip]{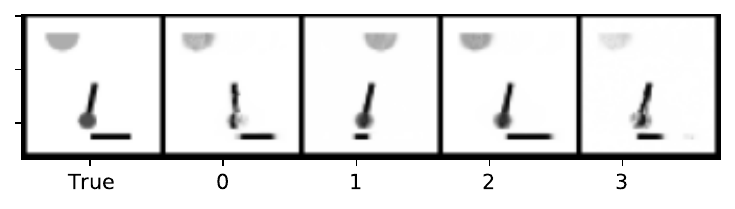}
    \caption{Interventions in the label-to-image VAE. Note that shadows respond to pendulum angle and sun position. }
    \label{fig:l2i_inter}\vspace{-3mm}
\end{figure}

\begin{figure}
    \centering
    \includegraphics[width=0.4\textwidth,trim={1mm 8mm 1mm 3mm},clip]{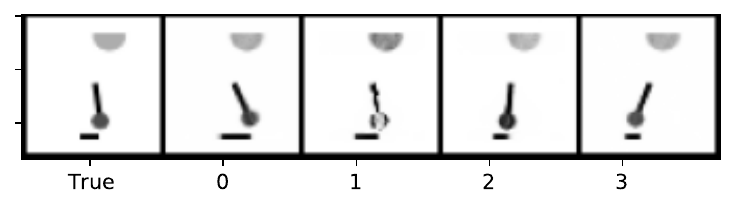}
    \includegraphics[width=0.4\textwidth,trim={1mm 3mm 1mm 3mm},clip]{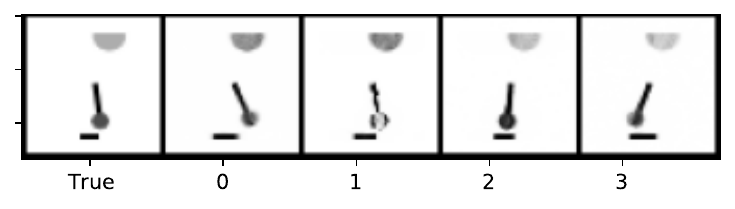}
    \caption{Label-to-image (Top) Image response to a sweep of the pendulum angle. Notice that for all interventions, the shadow responds to the pendulum. (Bottom) Image response after shadow position is debiased from pendulum angle. In this case, both the first and last intervened images show noticeable differences from the pre-counterfactual images and one can observe that the shadow midpoint remains consistent across the row. }
    \label{fig:l2i_sweep_pendulum}\vspace{-3mm}
\end{figure}

\begin{figure}
    \centering
    \includegraphics[width=0.4\textwidth,trim={1mm 8mm 1mm 3mm},clip]{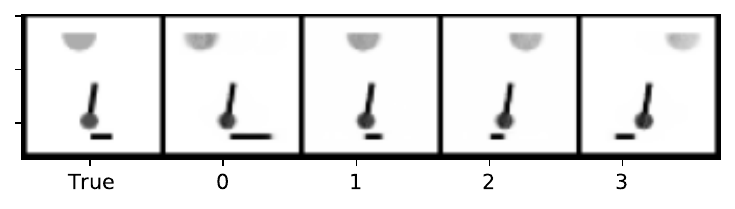}
    \includegraphics[width=0.4\textwidth,trim={1mm 3mm 1mm 3mm},clip]{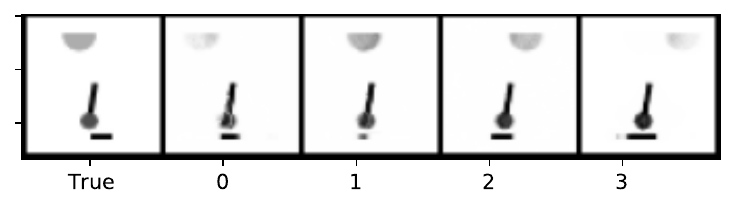}
    \caption{Label-to-image (Top) Image response to a sweep of sun positions. Again, notice that the shadow responds to the sun's position. (Bottom) image response after breaking the sun position to shadow position link. }
    \vspace{-2mm}
    \label{fig:l2i_sweep_sun}\vspace{0mm}
\end{figure}

Other than this additional pre-training step to learn the $\mathbf{A}$ matrix, the encoding step and the causal layer steps are still operating exactly the same as in the label-to-label VAE. We simply attach an image decoder after the causal layer. As in \cite{causalvae}, we see that the images have mostly disentangled the endogenous variables and the encoder is able to create an image where interventions can happen. These images are displayed in Figure \ref{fig:l2i_inter}. Notice especially in the shadow position intervention that the sun position and pendulum angle have not changed. With \alg{}, we can recreate the results from the label-to-label VAE in the label-to-image generator. These results are shown in Figures \ref{fig:l2i_sweep_pendulum} and \ref{fig:l2i_sweep_sun}. 

As in the label-to-label space, the top row of both of the figures shows the sweeps of $\theta$ and $x_{sun}$ in their latent space. With the interventions, the shadow responds accordingly and the complementary exogenous variable stays relatively consistent. The bottom row shows that the shadow position no longer responds to the interventions that are being enforced. In both Figure \ref{fig:l2i_sweep_pendulum} and \ref{fig:l2i_sweep_sun}, the first and the last interventional images show noticeable movement from the non-counterfactual interventions. 




\subsection{Mindset Data}

\begin{figure}
    \vspace{-1mm}
    \centering
    \includegraphics[width=0.43\textwidth,trim={1.2cm 6mm 1.2cm 1.2cm},clip]{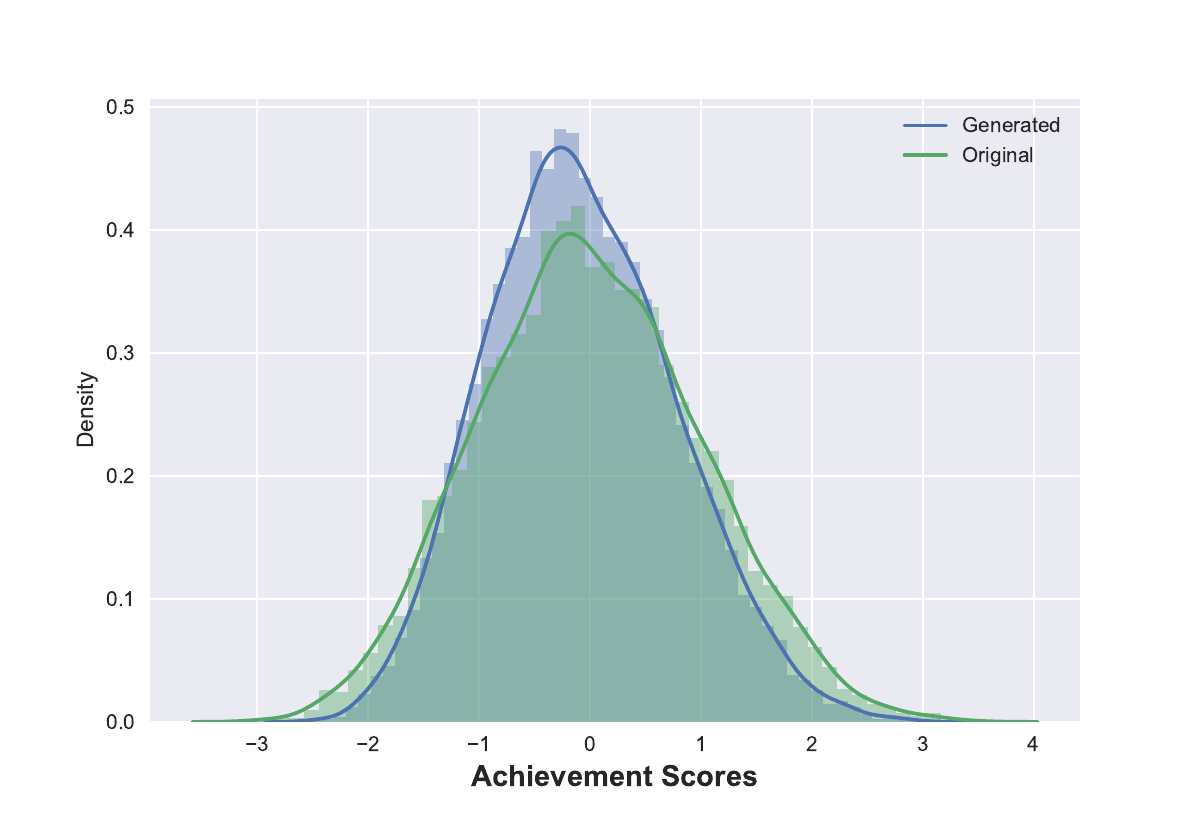}
    \caption{Generated distribution of achievement scores closely matches the distribution in the original dataset.}
    \label{fig:mindset_ach_recon}\vspace{-3mm}
\end{figure}

We begin by first training our \alg{} method on the original Student Mindset data. In Figure \ref{fig:mindset_ach_recon}, we see the generated achievement score from our model compared to the original dataset. The model is able to regenerate each of the feature distributions. We take the top 30\% as having the intervention (1) and lower 70\% as no intervention (0) since this matches the base rates in the original dataset. The current \alg{} operates on continuous variables and we treat the intervention values generated as a probability of treatment. 

The results of an ATE on the generated dataset vs the three baseline measurements on the original dataset of ATE, IPW, and AIPW are shown in Figure \ref{fig:mindset_bootstrap}. The generated dataset overestimates the ATE bias. We intervene on the path from student expectation to achievement scores, breaking the strongest positive correlation and see the \alg{} ``De-Bias" ATE estimation clearly drop below the advanced adjustment baseline methods. This makes intuitive sense since we leave the negatively correlated school mindset confounder which likely plays a small role in underestimating the ATE. Note that the advanced methods themselves are not necessarily ground truth, but they reflect an approximate ATE we would expect a ``de-biased" dataset to have. 

\begin{figure}
    \centering
    \includegraphics[width=0.45\textwidth,trim={1.2cm 6mm 1.2cm 1.2cm},clip]{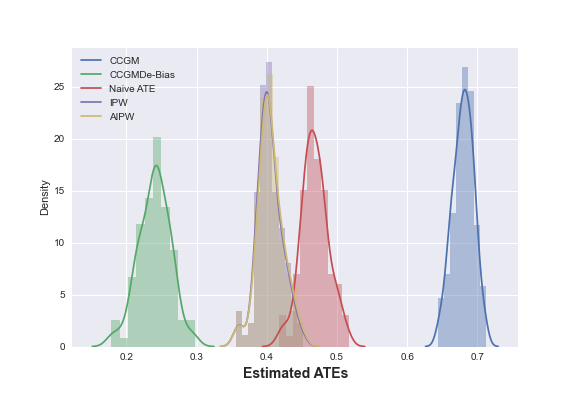}
    \caption{Bootstrap distribution of ATEs for various methods and baselines sampling the entire dataset (n=10391) 100 times.}
    \label{fig:mindset_bootstrap}\vspace{-3mm}
\end{figure}

Further, Table \ref{table:ATEs} shows the full results for all models and multiple de-biasing schemes. It is clear that our generative model is able to produce a "de-biased" dataset accounting for the positive ATE bias for the Student Expectation confounder. Although removing both confounders does marginally increase the 95\% bounds for out ATE as we would expect for a negative bias, it does not also do so when we only remove the school mindset. Here it becomes clear our model puts very little weight on this negative bias, a possible limitation of our model with the noisy nature of this dataset and a point of interest for further inquiry. All results are shown with empirical standard deviations and 95\% confidence interval bounds using bootstrap sampling methods ($iters=100$). 

\begin{table}[!t]
\centering
\renewcommand{\arraystretch}{1.3}
\newcommand*\rot{\rotatebox{0}}
\caption{Mindset ATE Results}
\begin{tabular}{l||>{\bfseries}rrrr}
\hline
{} &  Mean ATE &  Std Dev &  [.025 &  .975] \\
\hline
\hline
\scriptsize CCGM                             &     0.680 &    0.015 &  0.650 &  0.707 \\ \hline
\scriptsize CCGM De-Bias \\ \scriptsize Student Expectation &     0.240 &    0.023 &  0.192 &  0.283 \\ \hline
\scriptsize CCGM De-Bias \\ \scriptsize Both                &     0.242 &    0.022 &  0.204 &  0.282 \\ \hline
\scriptsize CCGM De-Bias \\ \scriptsize School Mindset      &     0.677 &    0.017 &  0.647 &  0.712 \\ \hline
\scriptsize AIPW                             &     0.405 &    0.018 &  0.364 &  0.441 \\ \hline
\scriptsize IPW                              &     0.404 &    0.018 &  0.363 &  0.440 \\ \hline
\scriptsize Naive                            &     0.468 &    0.019 &  0.426 &  0.507 \\ \hline
\end{tabular}
\label{table:ATEs}\vspace{-3mm}
\end{table}
\section{Conclusion}

In this paper, we demonstrate the value of \alg{}, an extension of previous causal generative model work that allows greater flexibility when considering counterfactual models and generating "out of distribution" data. We demonstrate the benefits of such a model on a simulated physics image dataset. We show the range of interventions and simulation of images outside of the training data, and outside of the ground-truth physics, with a simple adjustment after training the model. We then demonstrate results on a tabular dataset where ground-truth is not known. We show that we can learn the original data distributions, and simulate datasets which remove the impact of confounders in ways the make intuitive sense based on advanced statistical adjustment baselines. Much work is needed on refining the precision on noisy datasets, extending the framework to more complex causal models, and exploring the limitations based on the noise present and the target causal structure if known. We believe \alg{} is a promising start within a growing field of work in causal generative models. 
\bibliographystyle{IEEEtran}
\bibliography{IEEEabrv,causal-debias}
\newpage
\onecolumn
\appendix
\section*{Other Points about the Latent-to-Latent VAE}
\begin{figure}[h]
    \centering
    \includegraphics[width=0.5\textwidth]{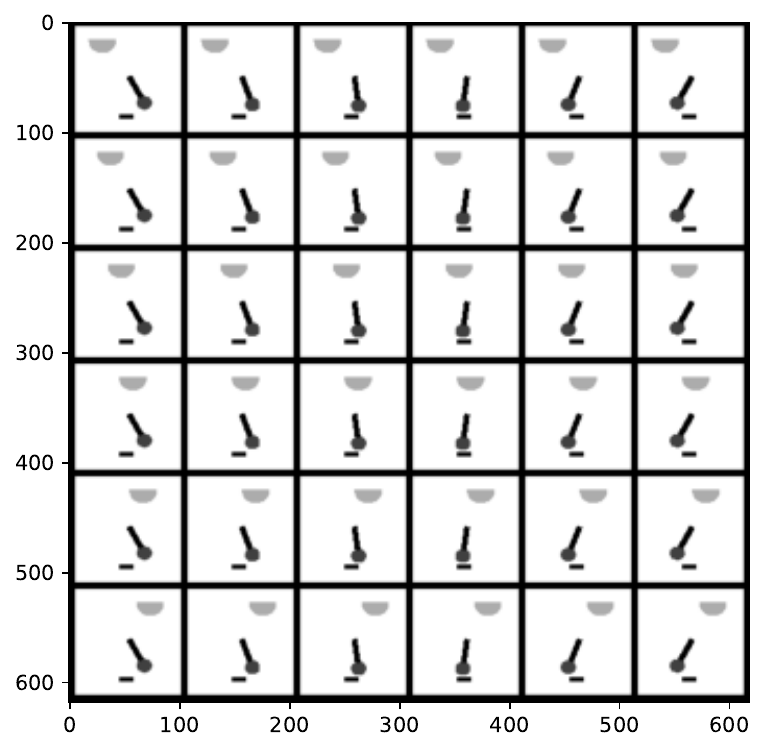}
    \includegraphics[width=0.5\textwidth]{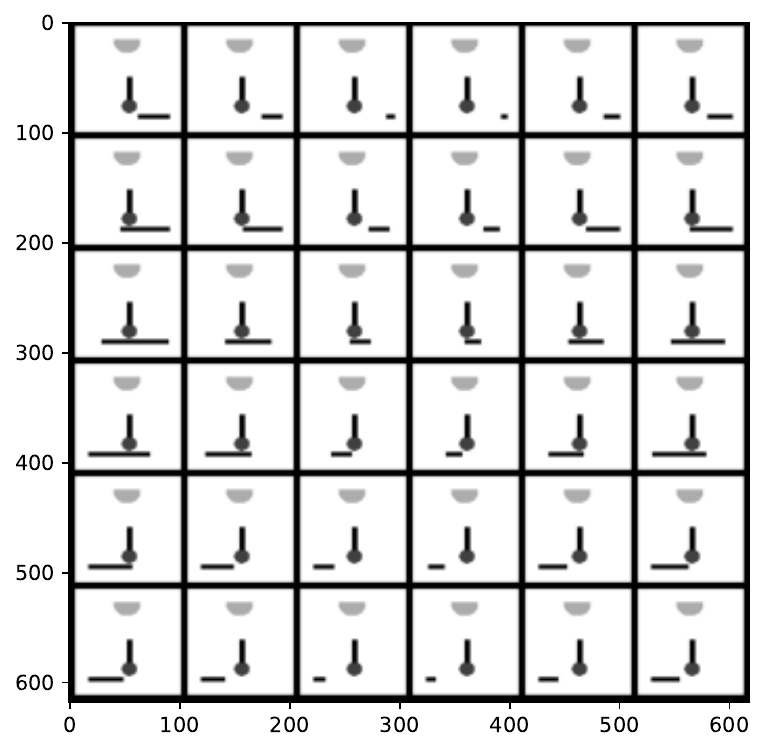}
    \caption{Visualization of the latent space of the causal concepts}
    \label{fig:l2l_latentviz}
\end{figure}


\newpage
\section*{Tabular Variable Reconstructions}

\begin{figure}[h]
    \centering
    \includegraphics[width=0.5\textwidth]{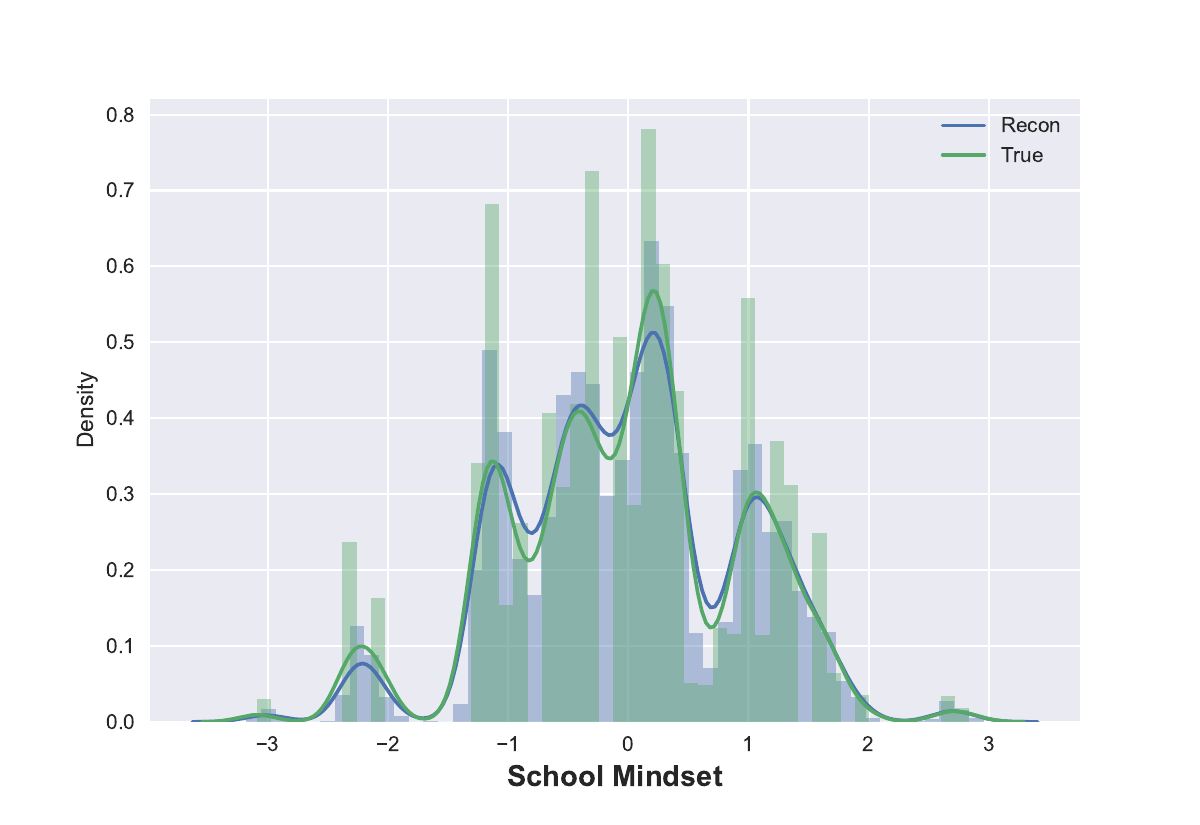}
    \caption{School mindset distribution CCGM reconstruction example}
    \label{fig:mindset_bootstrap}
\end{figure}

\begin{figure}[h]
    \centering
    \includegraphics[width=0.5\textwidth]{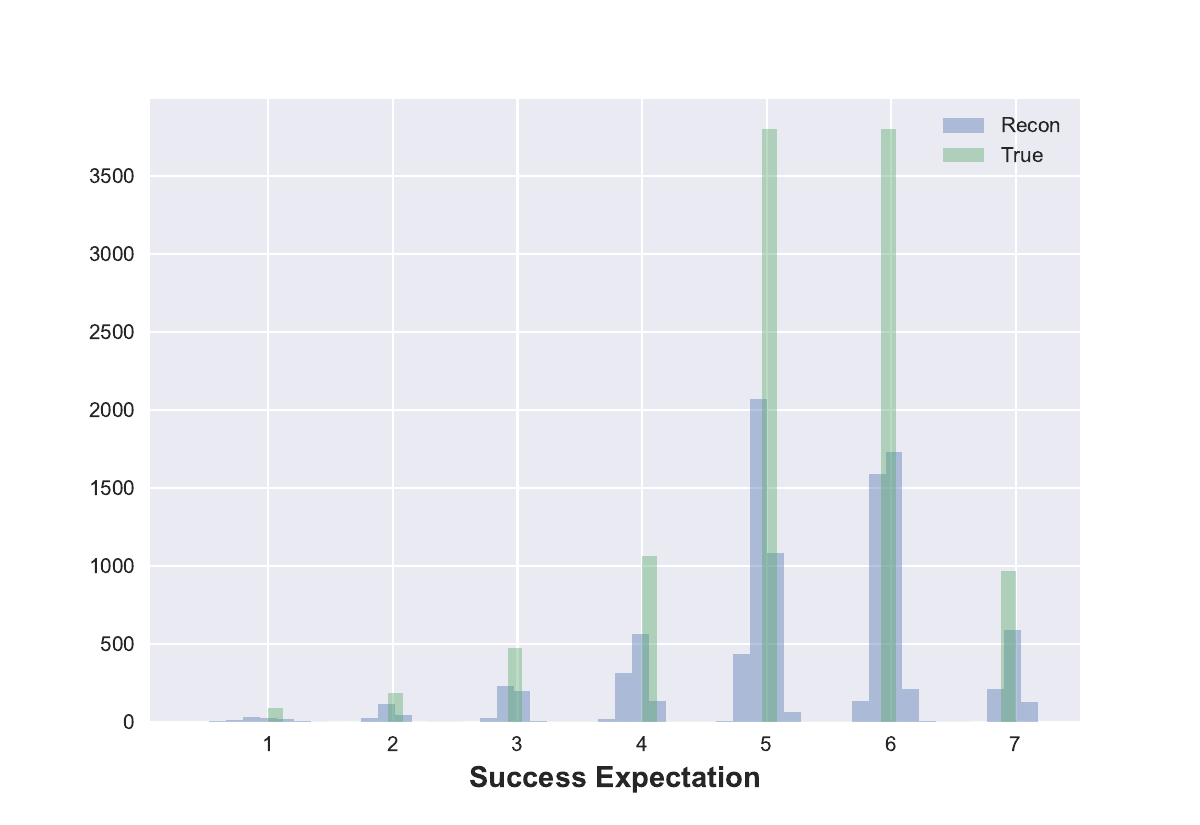}
    \caption{Success Expectation distribution CCGM reconstruction example}
    \label{fig:mindset_bootstrap}
\end{figure}

\begin{figure}[h]
    \centering
    \includegraphics[width=0.5\textwidth]{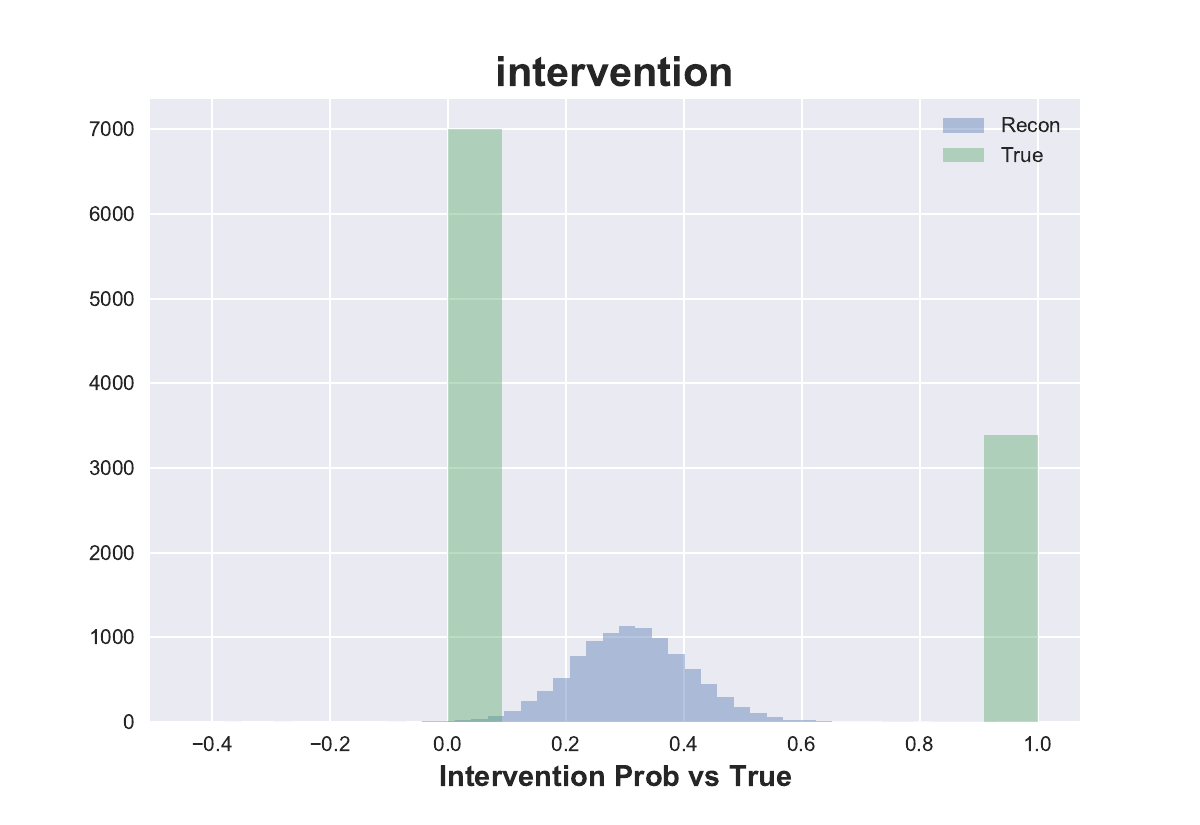}
    \caption{Intervention probability distribution CCGM reconstruction example. Note that top 30\% is used to bin data into binary before calculating ATEs.}
    \label{fig:mindset_bootstrap}
\end{figure}

\end{document}